\documentclass[12pt]{article}
\usepackage[a4paper]{geometry}
\usepackage[font=footnotesize,labelfont=bf]{caption}
\usepackage{fancyhdr}
\usepackage{subcaption}
\usepackage{placeins}
\usepackage{amsmath}
\usepackage{amsfonts}
\usepackage{graphicx,psfrag,epsf}
\usepackage{enumerate}
\usepackage[linesnumbered,ruled]{algorithm2e}
\usepackage{tabu}
\usepackage{setspace}
\usepackage{tikz}
\usepackage{float}
\usepackage{bm}
\usepackage{color}
\usepackage{listings}
\usepackage[square,sort,comma,numbers]{natbib}
\usepackage[hidelinks]{hyperref}
\usepackage[open]{bookmark}
\usepackage{booktabs}
\usepackage{rotating}
\usepackage{authblk}

\SetArgSty{textnormal}
\SetKwInput{KwTrain}{Train}

\fancypagestyle{firststyle}
{
	\fancyhf{}
	\fancyfoot[L]{\fontsize{9}{11}\selectfont Accepted for publication in \textit{Information Fusion} at \url{https://doi.org/10.1016/j.inffus.2024.102524}. \newline \copyright \ 2024. This manuscript version is made available under the CC-BY 4.0 license \url{http://creativecommons.org/licenses/by/4.0/}}
}

\begin{document}

\title{Stacked Penalized Logistic Regression for Selecting Views in Multi-View Learning}
\date{June 7, 2024}
\author[1]{Wouter van Loon}
\author[1]{Marjolein Fokkema}
\author[1,2]{Frank de Vos}
\author[3]{Marisa Koini}
\author[3]{Reinhold Schmidt}
\author[1,2]{Mark de Rooij}
\affil[1]{Department of Methodology and Statistics, Leiden University}
\affil[2]{Leiden Institute for Brain and Cognition}
\affil[3]{Division of Neurogeriatrics, Department of Neurology, Medical University of Graz}	
\maketitle

\thispagestyle{firststyle}

\begin{abstract}
	\noindent Data for which a set of objects is described by multiple distinct feature sets (called views) is known as multi-view data. When missing values occur in multi-view data, all features in a view are likely to be missing simultaneously. This may lead to very large quantities of missing data which, especially when combined with high-dimensionality, can make the application of conditional imputation methods computationally infeasible. However, the multi-view structure could be leveraged to reduce the complexity and computational load of imputation. We introduce a new imputation method based on the existing stacked penalized logistic regression (StaPLR) algorithm for multi-view learning. It performs imputation in a dimension-reduced space to address computational challenges inherent to the multi-view context. We compare the performance of the new imputation method with several existing imputation algorithms in simulated data sets and a real data application. The results show that the new imputation method leads to competitive results at a much lower computational cost, and makes the use of advanced imputation algorithms such as missForest and predictive mean matching possible in settings where they would otherwise be computationally infeasible.			
\end{abstract}

\textbf{keywords} \textit{missing data}, \textit{imputation}, \textit{multi-view learning}, \textit{stacked generalization}, \textit{feature selection}

\newpage

\section{Introduction}

Multi-view data refers to any data set where the features have been divided into distinct feature sets \citep{multiview_review, multiview_book, Smilde2022}\footnote{Depending on the research area, multi-view data is sometimes called multi-block, multi-set, multi-group, or multi-table data \citep{Smilde2022}.}. Such data sets are particularly common in the biomedical domain where these feature sets, commonly called \textit{views}, often correspond to different data sources or modalities \citep{multiview_bio, UKbiobank, UKbiobank_imaging, ADNI}. Classification models of disease using information from multiple views generally lead to better performance than models using only a single view \citep{Schouten2016, deVos2016, deVos2017, Salvador2019, Guggenmos2020, Ali2021}. Traditionally, information from different views is often combined using simple \textit{feature concatenation}, where the features corresponding to different views are simply aggregated into a single feature matrix, so that traditional machine learning methods can be deployed \citep{multiview_bio}. More recently, dedicated multi-view machine learning techniques have been developed, which are specifically designed to handle the multi-view structure of the data \citep{multiview_book, multiview_bio}. One such multi-view learning technique is stacked penalized logistic regression (StaPLR) \citep{StaPLR}. In addition to improving classification performance, StaPLR can automatically select the views that are most relevant for prediction \citep{StaPLR, StaPLR2, StaPLR3}. This ability to select the most relevant views is particularly important in the biomedical sciences \citep{multiview_bio} where selecting, for example, a subset of brain scan types \citep{StaPLR3}, could drastically reduce costs in future measurements, and prevent patients from undergoing unnecessary medical procedures. Furthermore, models which select views rather than individual features tend to be more interpretable \citep{StaPLR3}.  \par 
In practice, not all views may be observed for all subjects. When confronted with missing views, typical approaches are to remove any subjects with at least one missing value from the data set (called \textit{list-wise deletion} or \textit{complete case analysis} (CCA)), or to replace missing values by some substituted value, a process known as \textit{imputation}. In biomedical studies, a single view may consist of thousands or even millions of features. With the traditional approach of feature concatenation, in the presence of missing views, CCA leads to a massive loss of information, while imputation may be computationally infeasible. In this article we propose a new method for dealing with missing views, based on the StaPLR algorithm. We show how this method requires much less computation by imputing missing values in a dimension-reduced space, rather than in the original feature space. We compare our proposed imputation method with imputation methods applied in the original feature space. 

\section{Methods} \label{sect:methods}


Missing values are often divided into three categories: missing completely at random (MCAR), missing at random (MAR), or missing not at random (MNAR) \citep{Rubin1976, Stef2018}. Values are said to be MCAR if the causes of the missingness are unrelated to both missing and observed data \citep{Stef2018}. Examples include random machine failure, or missingness introduced by analyzing a random sub-sample of the data. If the missingness is not completely random but depends only on observed data, the missing values are said to be MAR \citep{Stef2018}. If the missingness instead depends on unobserved factors, the missing values are said to be MNAR \citep{Stef2018}. Here, we will focus on MCAR missing values. \par 
The simplest way of dealing with MCAR missing values is to discard observations with at least one missing value through complete case analysis. However, this approach is potentially very wasteful since a single missing value causes an entire observation to be removed from the data. CCA may therefore remove many more observed values from the data than the number of values initially missing, and drastically reduce the sample size, leading to increased variance and therefore less accurate predictions. \par
To prevent wasting observed data, missing values can be imputed. The simplest form of imputation is to replace each missing value with a constant. A very common choice is the unconditional mean of the feature, a procedure known as (unconditional) mean imputation (MI). If one is primarily interested in prediction, MI has some favorable properties: Its computational cost is extremely small, and it has been shown that MI is universally consistent for prediction even for MAR data, as long as the learning algorithm used is also universally consistent \citep{Josse2019}. Here \textit{consistent} means that, given an infinite amount of training data, the prediction function achieves the error rate of the best possible prediction function (i.e., the \textit{Bayes rate}), while \textit{universal} means that the procedure is consistent for all possible data distributions \citep{Josse2019}. However, MI is often criticized because it is known to distort the data distribution by attenuating existing correlations between the features, underestimating the variance, and causing bias in almost any estimate other than the mean \citep{Stef2018}. \par 
Many more sophisticated imputation methods have been developed. The literature on the imputation of missing values is vast, and we do not aim to give a complete overview here. However, most of the popular imputation methods can be grouped in a number of categories. The first such category consists of \textit{cold deck} imputation methods, which impute missing values using observed values from a different data set \citep{colddeck}. However, this requires suitable additional data to be available, which is often not the case. By contrast, \textit{hot deck}-style imputation methods \citep{hotdeck} are more generally applicable. For each observation with missing values, these imputation methods find one or several complete observations in the data which are most similar to the observation with missing values \citep{hotdeck}. The observed values of these cases, or some function thereof, are then used to impute the missing values of the incomplete case. The most popular example is imputation based on the \textit{k}-nearest neighbors (kNN) algorithm \citep{Dixon1979}. A different category of imputation methods is that of \textit{regression}-based imputation. This includes the state-of-the-art multiple imputation through chained equations (MICE) \citep{mice}. Another category is based on \textit{matrix factorization}, which includes Adaptive-Impute \citep{AdaptiveImpute}, and various other methods \citep{missMDA} based on, for example, principal component analysis (PCA) \citep{PCAimputation} or multiple factor analysis (MFA) \citep{MFAimputation}. More recently, \textit{tree}-based imputation methods such as missForest \citep{missForest} have become popular. Finally, there are \textit{deep learning} imputation methods which are generally based on \textit{auto-encoders}, such as multiple imputation with denoising autoencoders (MIDAS) \citep{MIDAS} or missing data importance-weighted autoencoder (MIWAE) \citep{MIWAE}, and/or based on \textit{generative adversarial networks}, such as generative adversarial imputation nets (GAIN) \citep{GAIN} or graph imputation neural networks (GINN) \citep{GINN}. Some of the most sophisticated imputation methods may combine ideas from several of the aforementioned categories. Predictive mean matching (PMM) \citep{Stef2018}, for example, uses regression-based imputation to find cases in the data which are most similar in terms of their predicted values. It is worth noting that it is generally preferable to generate not one, but multiple imputed data sets, so that correct variance estimates can be obtained \citep{Stef2018}; this is known as \textit{multiple imputation} \citep{Stef2018}. \par
We can also categorize the existing imputation methods depending on whether they perform \textit{unconditional} or \textit{conditional} imputation. We define an unconditional imputation method as any method in which the imputation of a missing value is based solely on other observations of the same feature, that is, the imputation takes place within a single column of the feature matrix. The aforementioned mean imputation is a classic example of an unconditional imputation method. By contrast, a conditional imputation method is any method in which the imputation of a missing value is based, in part or completely, on observations of other features, that is, the imputation uses different columns of the feature matrix. Most sophisticated imputation methods, such as Bayesian multiple imputation and PMM, are conditional imputation methods. The distinction between unconditional and conditional imputation methods is of particular interest for feature selection. Unconditional imputation methods, such as mean imputation, use only the univariate distributions for imputation, so that the imputed feature remains in some sense `pure' and free from contamination from other features. However, as mentioned earlier, mean imputation is known to distort the data distribution by attenuating existing correlations between the features \citep{Stef2018}. \par By contrast, some (but not all) conditional imputation methods preserve the correlations between features \citep{Stef2018}. However, in this case the imputed values depend on other features in the data. In the event that a selected feature has a large number of imputed values, this may lead to difficulties in interpretation, since a large proportion of the selected feature is derived from other features. Nevertheless, a recent study on the effect of imputation methods on feature selection suggests sophisticated conditional imputation methods generally lead to better results than unconditional imputation methods \citep{Mera2021}. Because it is not possible to both perform the imputation independent of other features and preserve existing correlations, one has to choose between one or the other. \par
It should be noted that other methods for handling missing data exist which do not explicitly impute missing values. These methods incorporate the missing data handling directly into the model fitting procedure and include likelihood-based methods such as full information maximum likelihood (FIML) \citep{Arbuckle1996, Myrtveit2001} for parametric regression models, and missingness incorporated in attributes (MIA) \citep{Twala2008} for decision trees. However, these methods are less broadly applicable than imputation methods \citep{Stef2018, Josse2019, Myrtveit2001} and we do not consider them here. \par

\subsection{From Missing Features to Missing Views}

In multi-view data, it is likely that missingness will occur at the view level, rather than at the feature level \citep{Song2020, Hornung2023}. Missing views may occur at random and/or by design \citep{Song2020, Hornung2023}. In a study where one of the views corresponds to features derived from a magnetic resonance imaging (MRI) scan, factors like the MRI scanner experiencing machine failure, a mistake in the scanning protocol by the researcher administering the scan, or a subject simply not making it to their appointment in time due to heavy traffic, would lead to all features of this view being simultaneously missing. Likewise, if one of the views corresponds to features derived from a sample of blood or cerebral spinal fluid (CSF), a lost or contaminated sample would lead to all derived features being simultaneously missing. Note that in these cases, although the missingness occurs at the view level, the underlying mechanism is still MCAR. Another common example of MCAR data occurs in the case of \textit{planned missingness}, where the missing values are part of the study design. For example, it may be considered too expensive to administer an MRI scan to all study participants, so instead an MRI scan is administered only to a random sub-sample of the participants. Again the underlying mechanism is MCAR, but all features corresponding to the MRI scan will be missing simultaneously for the unmeasured sub-sample. Throughout the rest of this article we will assume that (1) for each observation, a view is either completely missing or completely observed, and (2) the missingness is completely at random (MCAR). \par 
Conceptually, one could impute a missing view by first applying feature concatenation, and then simply applying a chosen imputation method on the concatenated feature set. However, in practice this may be impossible. For example, if the missing view is an MRI scan, there could be hundreds of thousands or even millions of missing values. Similar numbers of missing values may occur with views corresponding to other neuroimaging techniques, gene-expression arrays, or other omics data. With such vast amounts of missing values, application of sophisticated multiple-imputation methods may be computationally infeasible. For example, multiple imputation by chained equations, a very popular method for handling missing data implemented in the R package \texttt{mice} \citep{mice} involves iteratively cycling through a procedure of model-fitting and generating imputed values from the fitted model. Such iterative approaches quickly become prohibitive with large numbers of predictor variables, where the computational load of even a single imputation will exceed that of fitting the final predictive model of interest. This process is further complicated by the small sample sizes and thus extreme high-dimensionality commonly seen in neuroimaging, which means that even in instances where the imputation would be computationally feasible, the obtained results may be of poor quality (see \ref{sect:pmm} for further explanation). \par
Several methods specifically aimed at dealing with missing values in multi-view data have recently been proposed. We again focus only on imputation methods, although other ways for dealing with missing data exist which do not involve imputation \citep{Hornung2023, Darrell2012, MVAE}. Multi-view imputation methods are typically generalizations of the categories described in the previous section, including multi-view imputation methods based on kNN \citep{TOBMI, Xie2023}, regression \citep{Hieke2016, Klau2018, PrediXcan, FUSION, TIGAR}, matrix factorization \citep{Xu2015, Cai2016, Zhang2018, Thung2014, Thung2018, Linder2019, Zhu2020, Zhang2020, Gong2023, MOFA, MOFA+}, random forests \citep{Cao2023}, auto-encoders \citep{Tran2017, MMAE, PRIME, Wu2023}, and generative adversarial networks \citep{Cai2018, VIGAN, Dai2021}. For a detailed comparison between some of these methods we refer to two recent reviews on multi-view missing data handling by \citet{Song2020} and \citet{Hornung2023}. \par
However, it is important to note that many studies introducing new imputation methods focus on measures of reconstruction quality \citep[e.g.,][]{MIWAE, GAIN, GINN, Xu2015, Zhang2018, Gong2023, Tran2017,  MMAE, Wu2023, Cai2018}. This means that values are removed from a data set, the missing values are imputed, and then the imputed values are compared with the `true' values that were previously removed. The method which obtains the lowest root mean squared error (RMSE) or a similar measure is then considered to be the `best' \citep{MIWAE, GAIN, GINN, Xu2015, Zhang2018, Gong2023, Tran2017,  MMAE, Wu2023, Cai2018}. But imputation is not prediction: The goal of imputation is to find \textit{plausible} values for the missing data, not to reconstruct the missing values with perfect accuracy \citep{Stef2018}. In fact, bad imputation methods can obtain the lowest RMSE values, and in general any measures based on the similarity between the `true' and imputed values are not useful for comparing imputation methods \citep{Stef2018}. This can also be observed empirically in terms of prediction error: Large improvements in reconstruction quality often do not translate to substantial improvements in predictive performance \citep[see e.g.,][]{MIWAE, GAIN, GINN, MMAE}. This means the added value of these newer imputation methods over longer-standing methods is not clear. \par 
Many of the proposed methods are also highly specific to certain types of data. Imputation methods that were developed for multi-omics data, for example, are typically based on known biological relations between data types, and are often limited to settings with only 2 views \citep{Song2020, TOBMI,PrediXcan, FUSION, TIGAR}. Deep-learning based approaches to multi-view imputation often focus on image data \citep{Tran2017, MMAE, Wu2023, Cai2018}, and although they are not limited to such data by definition, changing the input data or the number of views often requires significant changes to the network architecture \citep{VIGAN}. Furthermore, deep learning-based imputation methods may require large sample sizes to perform well \citep{GAIN}, which is at odds with the extreme high-dimensionality that often occurs in neuroimaging or omics analysis. Many methods are also not (yet) available as public software. \par 	
In this article, we propose a general method for the imputation of missing views based on the StaPLR algorithm. The main advantages are (1) that it can be used for any type of tabular data, including high-dimensional data, (2) any existing imputation algorithm can be used without having to adapt it to the multi-view setting and (3) that the imputation takes place in a dimension-reduced space, drastically reducing computational load. We evaluate the proposed method on simulated and real data, using outcome measures of classification and view selection performance.

\subsection{Multi-View Stacking and the StaPLR Algorithm}

\FloatBarrier

\begin{sidewaysfigure}[h!]
	\centering
	\resizebox{.99\linewidth}{!}{\includegraphics{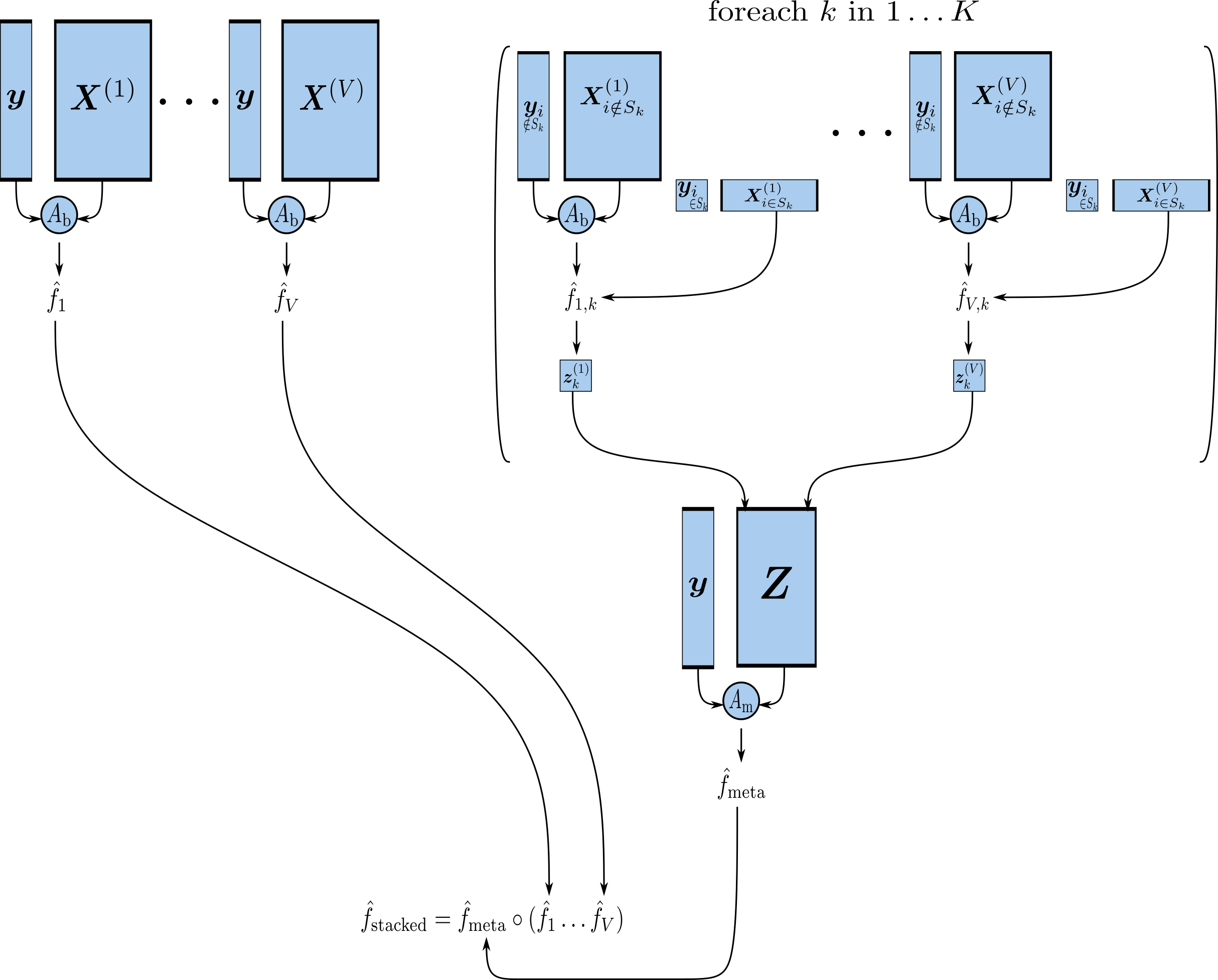}}
	\caption{The 2-level MVS algorithm, with a single base learner, represented as a flow diagram. StaPLR denotes the special case where all learners are penalized logistic regression learners. \label{fig:staplr_flow}}
\end{sidewaysfigure}

The general StaPLR algorithm was proposed by \citet{StaPLR}, and later extended to more than 2 levels of stacking \citep{StaPLR3}. We will briefly revisit it here. StaPLR is a form of multi-view stacking (MVS) \citep{StaPLR, Li2011, multiview_stacking} for binary classification problems, with specific choices for the learning algorithms. For clarity, we will assume throughout this article that (1) there are only 2 levels of stacking, the \textit{base} (feature) level, and the \textit{meta} (view) level, and (2) there is only one base learning algorithm ($A_\text{b}$), which is the same for all views. Under these assumptions the MVS algorithm can be given by Algorithm \ref{al:1}, and displayed as a flow diagram as in Figure \ref{fig:staplr_flow}. For each view $\bm{X}^{(v)}, v = 1 \dots V$, a trained classifier $\hat{f}_v$ is obtained by applying the base-learning algorithm $A_{\text{b}}$ to the pair $\bm{X}^{(v)}, \bm{y}$. For each of the base-classifiers $\hat{f}_v, v = 1 \dots V$, $k$-fold cross-validation is performed to obtain a vector $\bm{z}^{(v)}$ of estimated out-of-sample predictions. We assume that these predictions take the form of predicted probabilities rather than hard class labels. The vectors $\bm{z}^{(v)}, v = 1 \dots V$, are concatenated column-wise into the matrix $\bm{Z}$, which is then used together with outcome $\bm{y}$ to train the meta-learning algorithm $A_{\text{m}}$ and obtain the meta-classifier $\hat{f}_{\text{meta}}$. The final classifier is then given by $\hat{f}_{\text{stacked}} = \hat{f}_{\text{meta}} \circ (\hat{f}_{\text{1}} \dots \hat{f}_{\text{V}})$. \par
StaPLR denotes the special case of MVS where all learners are penalized logistic regression learners \citep{StaPLR}. In particular, we choose $A_{\text{b}}$ to be an $L_2$-penalized logistic regression learner including cross-validation for the tuning parameter, so that each $\hat{f}_v$ is a fully tuned logistic ridge regression classifier \citep{StaPLR}. For $A_{\text{m}}$ we choose an $L_1$-penalized nonnegative logistic regression learner including cross-validation for the tuning parameter, so that $\hat{f}_{\text{meta}}$ is a fully tuned logistic nonnegative lasso classifier \citep{StaPLR, StaPLR2}. In this setup, the base-learners induce shrinkage within each view, while the meta-learner selects or discards views based on their predictive relevance \citep{StaPLR}. StaPLR, and the MVS model in general, share some similarities with the recently proposed Mixture-of-Views (MoV) model \citep{MoV}. Both are based on the ``divide and conquer" paradigm of dividing a complex problem into simpler ones and then combining them \citep{MoV}. However, the MoV model is a multi-view generalization of the mixture-of-experts model (MoE) \citep{MoE1, MoE2, MoE3}, while StaPLR/MVS is a multi-view generalization of stacking \citep{Wolpert1992}. This means there are two important differences between the models. The first is that in the MoV model, the weights are assigned to each of the view-specific classifiers based on within-sample performance, while in StaPLR/MVS they are assigned based on estimated out-of-sample performance. Basing the weights on estimated out-of-sample performance is the core idea behind stacking, and can filter out bias and improve generalization performance \citep{Wolpert1992}. The second difference is that in the MoV model the goal is to find the most predictive set of views for each individual sample, while in StaPLR the goal is to find the most predictive set of views across all samples. In our case, only the second type of view selection is of interest as it has the potential to reduce data collection in future experiments.

\begin{algorithm} 
	\caption{The 2-level MVS algorithm, with a single base learner. StaPLR denotes the special case where all learners are penalized logistic regression learners.} \label{al:1}
	\DontPrintSemicolon
	\KwData{Views $\bm{X}^{(1)}, \dots, \bm{X}^{(V)}$ and outcomes $\bm{y} = (y_1, \dots, y_n)^T$.}
	\ForEach{$v$ = 1 to $V$}{
		$\hat{f}_{v} = A_{\text{b}}(\bm{X}^{(v)}, \bm{y})$ \label{al:1:basetrain} \;
		\ForEach{$k$ = 1 to $K$}{
			$\hat{f}_{v,k} = A_{\text{b}}(\bm{X}^{(v)}_{i \notin S_k}, \bm{y}^{}_{i \notin S_k})$  \label{al:1:cvtrain} \;
			$\bm{z}^{(v)}_{k} = \hat{f}_{v,k}(\bm{X}^{(v)}_{i \in S_k})$
		}
	}
	$\bm{Z} = (\bm{z}^{(1)}, \dots, \bm{z}^{(V)})$\;
	$\hat{f}_{\text{meta}} = A_{\text{m}}(\bm{Z}, \bm{y})$\;
	$\hat{f}_{\text{stacked}} = \hat{f}_{\text{meta}} \circ (\hat{f}_{\text{1}} \dots \hat{f}_{\text{V}})$
\end{algorithm}

\FloatBarrier

\subsection{The Proposed Imputation Method}

\FloatBarrier

As shown in Algorithm \ref{al:1} and Figure \ref{fig:staplr_flow}, StaPLR performs view selection through two steps: a dimension reduction step, and a feature selection step. First, a matrix of cross-validated predictions $\bm{Z}$ is constructed, in which each view is represented by a single column. The meta-learning algorithm then performs feature selection in this reduced space, which translates to view selection in the original feature space. Note that the reduction of dimensionality is generally very large: If the complete concatenated feature matrix $\bm{X} = (\bm{X}^{(1)}, \dots, \bm{X}^{(V)})$ has dimensionality $n \times m$, then $\bm{Z}$ has dimensionality $n \times V$, and generally $V \ll m$. This further implies that, while the original data is usually very high-dimensional, this is often not the case in the reduced space; see for example \citet{StaPLR3} for an application of StaPLR where $n \ll m$, but $n > V$. Obviously, it is much less computationally intensive to fit models in the reduced space, than it is to fit them in the original space. 
We therefore propose to impute any missing values in this reduced space rather than in the original feature space. Because of the difference in dimensionality, this is computationally very attractive. Furthermore, as long as we assume $n > V$, imputation methods which would perform poorly in the high-dimensional feature space can instead be applied in the reduced space. \par
\begin{figure}[h!]
	\centering
	\resizebox{.9\linewidth}{!}{\includegraphics{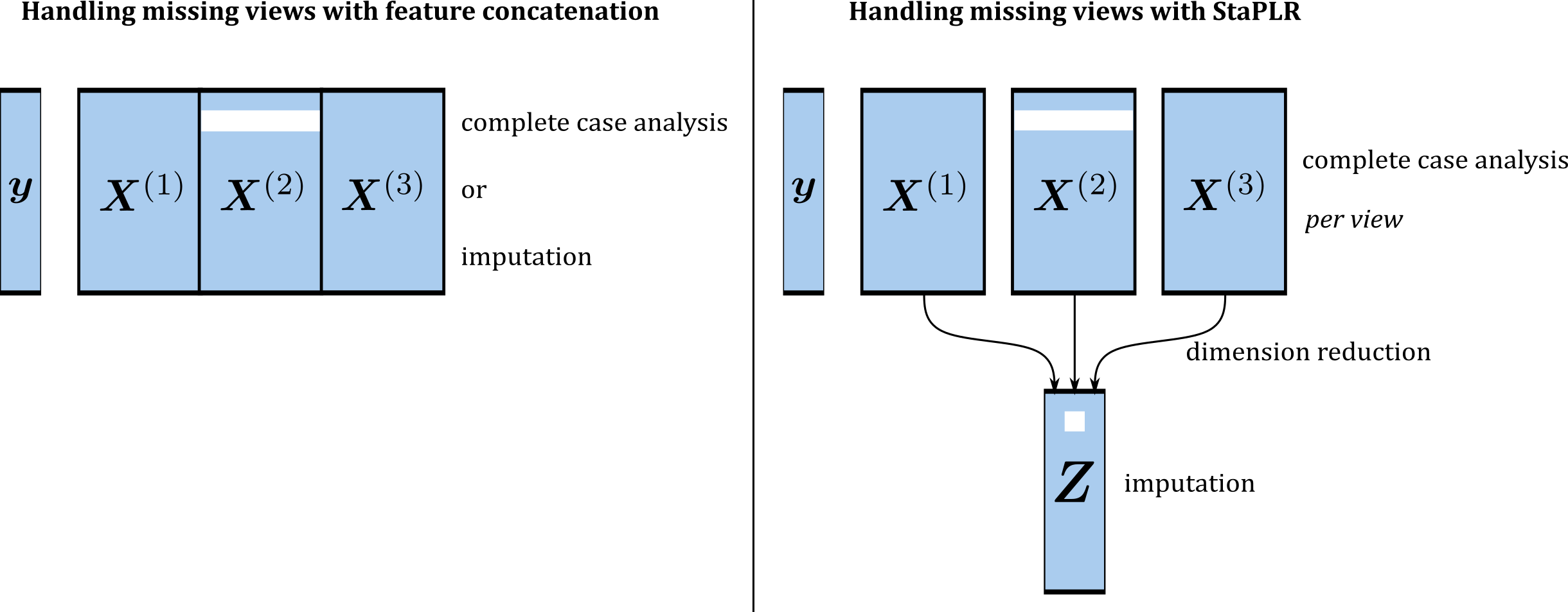}}
	\caption{An example of how missing views can be handled through either classical feature concatenation (left), or through StaPLR (right). The blank areas in the data represent missing values. In this example, we assume there are three views, namely $\bm{X}^{(1)}$, $\bm{X}^{(2)}$, and $\bm{X}^{(3)}$. We denote the number of features in each view by $m_1$, $m_2$, and $m_3$, respectively. Assume that there are $l$ observations which have missing values on $\bm{X}^{(2)}$. In the case of feature concatenation, we would have to either impute $lm_2$ missing values, or entirely discard $l(m_1 + m_3)$ observed values. However, in the proposed missing data handling method for StaPLR, only $l$ values need to be imputed. \label{fig:missing_data}}
\end{figure}

To apply the dimension reduction, the base-learning algorithm is trained on each view separately. Given that a view is either completely missing or completely observed, this means that within each application of the base learner there are no observations with partially observed data. Thus, the base learner can simply be trained on the set of complete cases for each view. A comparison of the proposed method for handling missing views in StaPLR and feature concatenation is shown in Figure \ref{fig:missing_data}. Note that although we use the StaPLR model specifically, the proposed method applies to multi-view stacking in general, and can most likely be easily extended to any form of late fusion.

\FloatBarrier

\section{Empirical Evaluation}


We evaluate the proposed meta-level imputation method by applying three popular imputation algorithms at the meta-level, and comparing the results with the application of those same algorithms at the feature level. We also compare the results with those of a dedicated multi-view imputation method. \par 
The first imputation algorithm is unconditional mean imputation (MI), where each missing value is replaced by the mean of the observed values of the feature.
The second algorithm is predictive mean matching (PMM) \citep{Stef2018}, a state-of-the-art multiple imputation algorithm implemented as the default in the R package \texttt{mice} \citep{mice}. PMM is based on Bayesian imputation under the normal linear model \citep{Stef2018, Rubin1987, Schafer1997}, but instead of using a Bayesian draw of the regression coefficient to directly calculate an imputed value, it is used to find a set of $d$ candidate `donors', which are the $d$ observations for which the predicted value under the imputation model is closest to the value drawn for the missing observation \citep{Stef2018, mice}. The imputed value is then a random draw from the \textit{observed} values of the donors. Advantages of PMM over direct Bayesian imputation include that the method allows for discrete target variables, the imputed values can never go outside the range of the observed data, and the method is more robust to model misspecification \citep{Stef2018}. Disadvantages compared to other imputation methods include the fact that PMM tends to work poorly in high-dimensional settings, and that the required computation time is so large that we can only apply PMM at the meta-level, as feature-level imputation using PMM proved entirely infeasible (see \ref{sect:pmm}). When applying PMM, we use the default value of 5 for both the number of candidate donors and the number of imputations. With multiple imputation comes the need to somehow combine the results of the different imputed data sets. We evaluate two strategies for combining the different imputed data sets: The first strategy is to generate the matrix of cross-validated predictions $\bm{Z}$ once, use the imputation model to generate 5 imputations, and then take the average of the imputed values to train the meta-learner. The second strategy is to generate the matrix $\bm{Z}$ 5 times, each time using different cross-validation folds, and then impute each matrix once before averaging the matrices. The benefit of the second strategy is that it also averages over different ways in which observations are allocated to cross-validation folds, but it is approximately 5 times more computationally expensive. To distinguish between the two strategies, we will refer to them as mPMM and cvPMM, respectively. \par 
The third algorithm we consider is missForest (MF), an iterative imputation method based on random forests \citep{missForest}, as implemented in the R package \texttt{missForest}. MF can be used to impute both continuous and categorical data, and by averaging over many unpruned regression or classification trees it is intrinsically a multiple imputation method, while producing only a single completed data set \citep{missForest}. When applying the MF algorithm, we use the default settings for the R package \texttt{missForest}, which constitutes growing 100 trees in each forest. \par 
We compare the results of the different imputation algorithms with those of a dedicated multi-view imputation algorithm based on multi-omics factor analysis (MOFA) \citep{MOFA,MOFA+}. As the name suggests, MOFA is an extension of factor analysis to the multi-view setting. Although developed for multi-omics data, it is more generally applicable because it only assumes that the features in each view follow a Gaussian, Poisson, or Bernoulli distribution \citep{MOFA,MOFA+}. MOFA decomposes the views into a shared factor matrix, and a set of view-specific weight matrices \citep{MOFA,MOFA+}, and is similar to multi-view learning with incomplete views (MVL-IV)\citep{Xu2015}. The matrix decomposition can then be used to impute missing values in the original views \citep{MOFA,MOFA+}. MOFA-based imputation is implemented in the R package \texttt{MOFA2} \citep{MOFA,MOFA+}, for which we use the default settings. \par
Finally, we also compare the results of the different imputation algorithms with complete case analysis (CCA), and with an application of StaPLR to the complete data before any missingness is generated, a process we will call \textit{complete data analysis} (CDA). Note that for all the conditional imputation methods the outcome $y$ is also included in the imputation procedure, as is generally recommended in the missing value literature \citep{Sterne2009}. For MOFA, we implemented this by including the outcome as an additional view following a Bernoulli distribution.

\subsection{Simulation Design}

We generate multi-view data consisting of $V = 4$ disjoint views. Each view $v = 1 \dots V$ consists of $m_v = 5 \times 10^{v - 1}$ random normally distributed features. We introduce a block correlation structure between the features: The mean correlation between features within the same view is set to $\rho_w$ = 0.5. The mean correlation between features in different views is set to $\rho_b$ = 0.2. If view $v$ corresponds to signal, then each feature within that view has a regression coefficient of $2/\sqrt{m_v}$ or $-2/\sqrt{m_v}$, each with probability 0.5. If a view corresponds to noise, then each feature within that view has a regression coefficient of zero. The features and their respective coefficients together form a linear predictor (with an intercept equal to zero) to which the logistic function is applied to obtain probabilities. These probabilities are then used to draw the observations of the binary outcome $y$. This method for generating simulated data is similar to that used in earlier work on StaPLR \citep{StaPLR, StaPLR2}. The following experimental factors are varied: 
\begin{itemize}
    \item Fraction of observations with a missing view: A single view is MCAR for either 50\% or 90\% of the observations.
    \item Size of the missing view: The missing view is either the smallest ($V_1$) or the largest ($V_4$) view. 
    \item Size of the noise view: Either the smallest ($V_1$) or the largest ($V_4$) view corresponds to noise, while all other views correspond to signal.
\end{itemize}
This leads to a total of $2 \times 2 \times 2 = 8$ experimental conditions. For each experimental condition, 100 replications are performed. For each replication of each condition, we generate a training set of $n = 1000$ observations in which the missingness occurs. All imputation and model fitting occurs in the training set. For each training set we generate a matching test set, also consisting of 1000 observations. There are no missing values in the test set, and the test set is only used for the calculation of certain outcome measures (Section \ref{sect:outcome_measures}).

\subsection{Software}

All simulations are performed in R (version 4.1.2) \citep{Rcore} on a high-performance computing cluster running Cent OS (Stream 8) with Slurm Workload Manager (version 20.11). All pseudo-random number generation is performed using the Mersenne Twister \citep{mersenne_twister}, R's default algorithm. Imputation with the missForest algorithm is performed using the R package \texttt{missForest} (version 1.5) \citep{missForest}, while predictive mean matching is performed using \texttt{mice} (version 3.14) \citep{mice}. All StaPLR model fitting is performed using \texttt{mvs} (version 1.0.2) \citep{mvs}, with a \texttt{glmnet} (version 4.1-4) \citep{glmnet} back-end. MOFA-based imputation was performed using \texttt{MOFA2} (version 1.8.0) \citep{MOFA,MOFA+}, with a \texttt{basilisk} (version 1.10.2) \citep{basilisk} back-end. Additional scripts, such as those used for mean imputation and data simulation, are included in the supplementary materials.  

\subsection{Outcome Measures} \label{sect:outcome_measures}

We compare the predictive performance of the trained stacked classifiers in terms of two outcome measures calculated on the test set. The first outcome measure is \textit{test accuracy}, which we define as the proportion of correctly classified observations in the test set given a classification threshold of 0.5. The second outcome measure is the \textit{mean squared error of probabilities} (MSEP), which we define as:
\begin{equation*}
    \text{MSEP} = \frac{1}{n}\sum_{i=1}^{n} (\hat{p}_i - p_i)^2, 
\end{equation*}
where $p_i$ is the true class probability of test set observation $i$, and $\hat{p}_i$ is the value predicted by the classifier. Note that outside of simulation studies the true class probabilities would be unknown, and the MSEP cannot be calculated. However, replacing the true probabilities in the MSEP with the true class labels naturally leads to a well-known classification measure known as the Brier score \citep{Brier1950}. Here we use the MSEP instead of the Brier score, as it is expected to be more sensitive to small changes in the data. \par
To assess view selection performance, we calculate four outcome measures. The first is simply the mean proportion of correctly selected views. The second outcome measure is the \textit{true positive rate} (TPR), which is defined as the average proportion of views truly related to the outcome that were correctly selected by the meta-learner. The third outcome measure is the \textit{false positive rate} (FPR), which is defined as the average proportion of views not related to the outcome that were incorrectly selected by the meta-learner. The fourth outcome measure is the \textit{false discovery rate}, which is defined as the average proportion of incorrectly selected views among the set of selected views if at least one view is selected, and zero otherwise. In addition to classification and view selection performance, we also calculate the logarithm of the computation time in seconds for each procedure. 

\section{Results}

\begin{figure}[h!]
	\centering
	\resizebox{.99\linewidth}{!}{\includegraphics{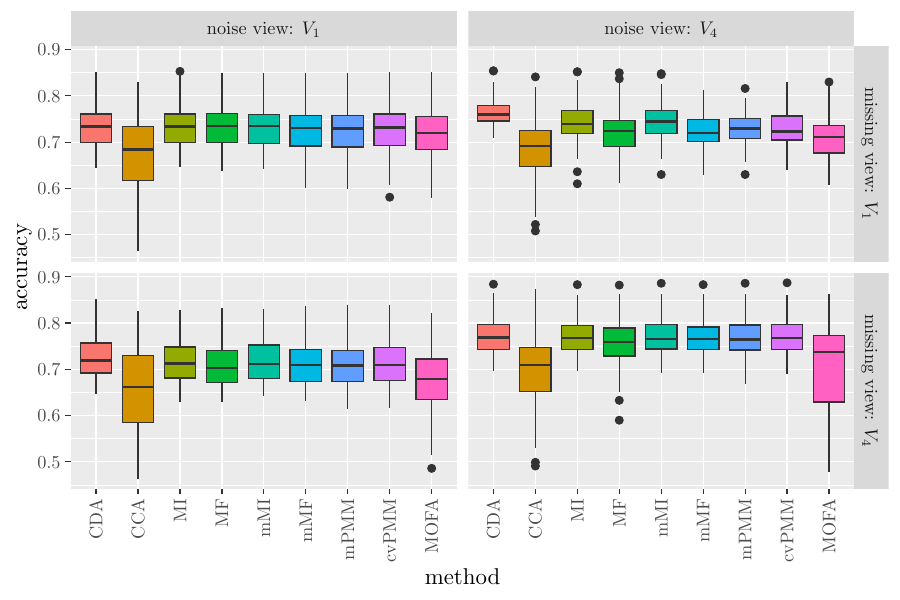}}
	\caption{Test accuracy with 90\% missingness. CDA = complete data analysis; CCA = complete case analysis; MI = feature-level mean imputation; MF = feature-level missForest; mMI = meta-level mean imputation; mMF = meta-level missForest; mPMM = meta-level predictive mean matching, generating $\bm{Z}$ once; cvPMM = meta-level predictive mean matching, generating $\bm{Z}$ five times; MOFA = multi-factor omics analysis imputation. $V_1$ is the smallest view, consisting of 5 features. $V_4$ is the largest view, consisting of 5000 features. \label{fig:acc_90}}
\end{figure}
\begin{figure}[h!]
	\centering
	\resizebox{.99\linewidth}{!}{\includegraphics{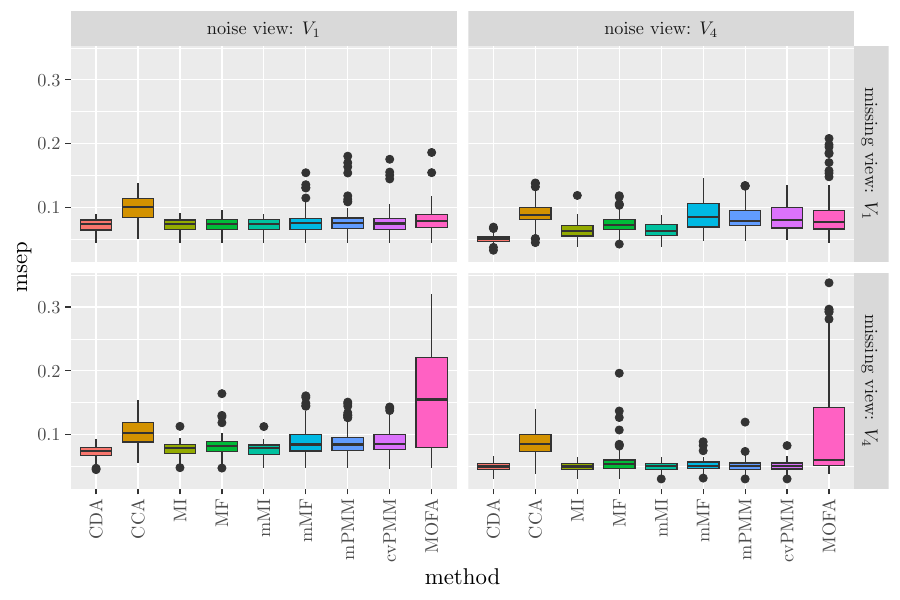}}
	\caption{Mean squared error of probabilities (MSEP) with 90\% missingness. CDA = complete data analysis; CCA = complete case analysis; MI = feature-level mean imputation; MF = feature-level missForest; mMI = meta-level mean imputation; mMF = meta-level missForest; mPMM = meta-level predictive mean matching, generating $\bm{Z}$ once; cvPMM = meta-level predictive mean matching, generating $\bm{Z}$ five times; MOFA = multi-factor omics analysis imputation. $V_1$ is the smallest view, consisting of 5 features. $V_4$ is the largest view, consisting of 5000 features. \label{fig:msep_90}}
\end{figure}
\begin{figure}[h!]
	\centering
	\resizebox{.99\linewidth}{!}{\includegraphics{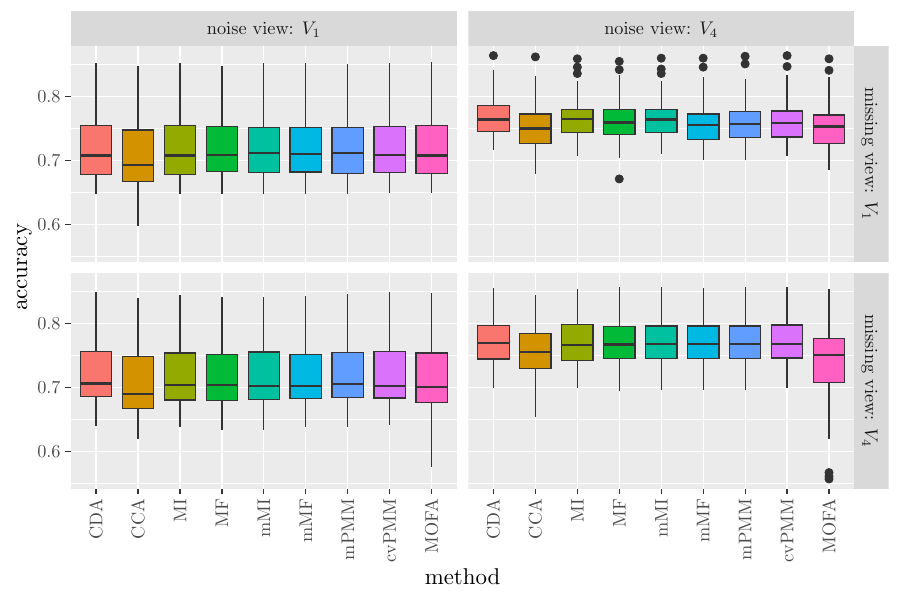}}
	\caption{Test accuracy with 50\% missingness. CDA = complete data analysis; CCA = complete case analysis; MI = feature-level mean imputation; MF = feature-level missForest; mMI = meta-level mean imputation; mMF = meta-level missForest; mPMM = meta-level predictive mean matching, generating $\bm{Z}$ once; cvPMM = meta-level predictive mean matching, generating $\bm{Z}$ five times; MOFA = multi-factor omics analysis imputation. $V_1$ is the smallest view, consisting of 5 features. $V_4$ is the largest view, consisting of 5000 features. \label{fig:acc_50}}
\end{figure}
\begin{figure}[h!]
	\centering
	\resizebox{.99\linewidth}{!}{\includegraphics{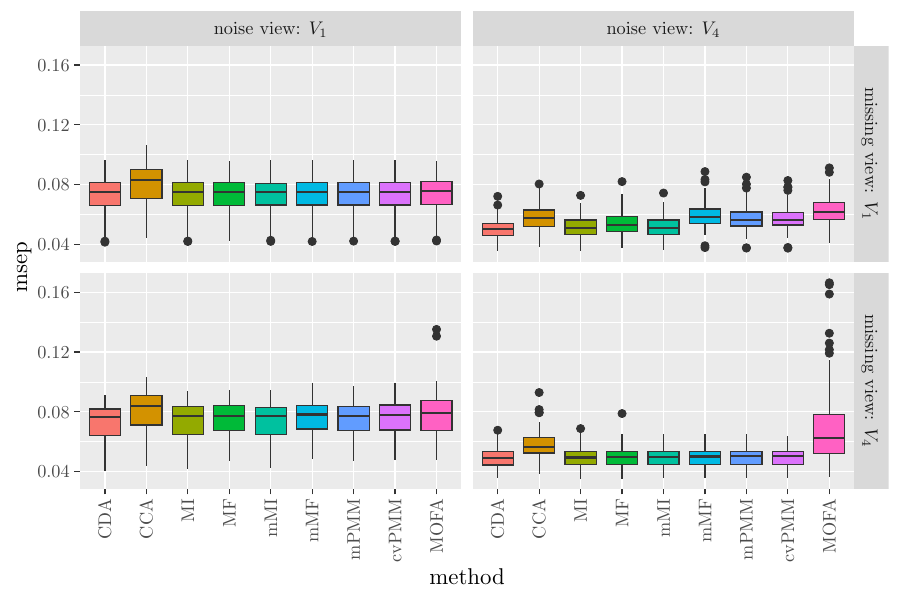}}
	\caption{Mean squared error of probabilities (MSEP) with 50\% missingness. CDA = complete data analysis; CCA = complete case analysis; MI = feature-level mean imputation; MF = feature-level missForest; mMI = meta-level mean imputation; mMF = meta-level missForest; mPMM = meta-level predictive mean matching, generating $\bm{Z}$ once; cvPMM = meta-level predictive mean matching, generating $\bm{Z}$ five times; MOFA = multi-factor omics analysis imputation. $V_1$ is the smallest view, consisting of 5 features. $V_4$ is the largest view, consisting of 5000 features. \label{fig:msep_50}}
\end{figure}

Figures \ref{fig:acc_90} and \ref{fig:msep_90} show, respectively, the classification accuracy and MSEP of the different methods for handling missing data under 90\% missingness. Figures \ref{fig:acc_50} and \ref{fig:msep_50} show the same outcome measures under 50\% missingness. We will focus on the results under 90\% missingness as they generally show the same pattern of performance for the different imputation methods, but in an emphasized manner. In terms of test accuracy, complete case analysis performs worse than all imputation methods. Among the imputation methods, MOFA imputation performs worst, particularly when the missing view consist of many features. The differences between the other imputation methods are comparatively small, although when the noise view consists of many features, and the smallest view ($V_1$) is missing, mean imputation appears to perform slightly better than the other imputation methods. \par
As expected, the MSEP is more sensitive to changes in the imputed data than the test accuracy. Again, MOFA performs particularly poorly when the missing view consists of many features. Excluding MOFA, CCA performs worst, while CDA performs best. Mean imputation performs as well as, or better than, the other imputation methods, and typically has a lower variance. Base-level missForest (MF) performs as well as, or better than, meta-level missForest (mMF) or predictive mean matching (mPMM and cvPMM). \par
For the sake of completeness, we also calculated the binomial deviance. It supports the same general conclusions, lying in between the accuracy and MSEP in terms of sensitivity to the imputed data (see \ref{sect:bin_dev}).

\FloatBarrier

\subsection{View Selection} \label{sect:view_selection}

In terms of the mean proportion of correctly selected views (Table \ref{tab:view_select}), CCA again performs worse than all imputation methods, except for one condition where MOFA imputation performs worst (Table \ref{tab:view_select}, row 3). A trend can be observed where mean imputation (both MI and mMI) performs well when the view to be imputed corresponds to noise, but performs poorly when the view to be imputed corresponds to signal. The mMF, mPMM and cvPMM methods all appear to perform similarly. mMF sometimes shows somewhat better performance than MF, particularly when the missing view has a large number of features, but the pattern of differences between the two methods is not entirely consistent. \par 
In terms of the true positive rate (Table \ref{tab:tpr}), CCA performs worse than all imputation methods. Both MI and mMI perform well in terms of TPR when compared with the other imputation methods. The mMF, mPMM and cvPMM method again show similar performance. The observed differences between MF and mMF do not appear to follow a consistent pattern. \par 
In terms of the false positive rate (Table \ref{tab:fpr}), a pattern can be observed where mean imputation (MI and mMI) performs really well when the view to be imputed corresponds to noise, but performs poorly when the view to be imputed corresponds to signal. The performance of MF shows the opposite behavior: It performs better when the view to be imputed corresponds to signal than when it corresponds to noise. Similar behavior can be observed for its meta-level implementation mMF. MOFA imputation also shows this pattern but in an extreme fashion: When the missing view corresponds to noise the FPR rises dramatically, up to 0.90. Comparing MF and mMF shows that MF has a slightly lower FPR when the size of the missing view is 5, but has a considerably higher FPR than mMF when the size of the missing view is 5000. The observed differences in FPR between mMF and mPMM and cvPMM are small under 50\% missingness. Under 90\% missingness larger differences are observed when the signal view with 5 features is missing, and when the noise view with 5000 features is missing. In terms of FPR, CCA performs well when compared to the imputation methods. \par 
The relative performance of the imputation methods in terms of false discovery rate (Table \ref{tab:fdr}) is similar to that in terms of FPR. CCA also performs fairly well in terms of FDR, just like it does in terms of FPR.

\FloatBarrier

\begin{table}
	\centering
	\setlength{\tabcolsep}{3pt}
	\resizebox{\textwidth}{!}{
		\begin{tabular}{lllrrrrrrrrr}
			\hline
			missingness & size of missing view & missing view is & CDA & CCA & MI & MF & mMI & mMF & mPMM & cvPMM & MOFA \\
			\hline
			50\% & 5 & noise & .903 & .850 & .900 & .903 & \textbf{.908} & .898 & .898 & .905 & .883 \\
			&  & signal & .960 & .918 & .943 & \textbf{.960} & .925 & .953 & .953 & .955 & .950\\
			& 5000 & noise & .958 & .908 & .955 & .910 & \textbf{.958} & .950 & .948 & .943 & .743\\
			&  & signal & .893 & .845 & .848 & .868 & .848 & .875 & .875 & .875 & \textbf{.935} \\
			\hline
			90\% & 5 & noise & .885 & .625 & \textbf{.895} & .865 & .880 & .858 & .855 & .873 & .768\\
			&  & signal & .970 & .703 & .868 & .873 & .875 & .895 & .908 & \textbf{.918} & .873 \\
			& 5000 & noise & .965 & .683 & \textbf{.983} & .870 & .978 & .930 & .953 & .963 & .750 \\
			&  & signal & .908 & .608 & .785 & .830 & .788 & .843 & .830 & .845 & \textbf{.850} \\
			\hline
	\end{tabular}}
	\caption{Mean proportion of correctly selected views. CDA = complete data analysis; CCA = complete case analysis; MI = feature-level mean imputation; MF = feature-level missForest; mMI = meta-level mean imputation; mMF = meta-level missForest; mPMM = meta-level predictive mean matching, generating $\bm{Z}$ once; cvPMM = meta-level predictive mean matching, generating $\bm{Z}$ five times; MOFA = multi-factor omics analysis imputation. The `best' (highest) values in each row are printed in bold (excluding the values for CDA and CCA since these are not imputation methods).}  \label{tab:view_select}
\end{table}

\begin{table}
	\centering
	\setlength{\tabcolsep}{3pt}
	\resizebox{\textwidth}{!}{
		\begin{tabular}{lllrrrrrrrrr}
			\hline
			missingness & size of missing view & missing view is & CDA & CCA & MI & MF & mMI & mMF & mPMM & cvPMM & MOFA \\
			\hline
			50\% & 5 & noise & .877 & .827 & .880 & \textbf{.890} & .887 & .887 & .887 & \textbf{.890} & .880\\
			&  & signal & .963 & .927 & \textbf{.970} & .950 & .963 & .943 & .943 & .943 & .937 \\
			& 5000 & noise & .957 & .900 & \textbf{.960} & \textbf{.960} & \textbf{.960} & \textbf{.960} & \textbf{.960} & .953 & .957 \\
			&  & signal & .877 & .810 & .853 & \textbf{.880} & .863 & .850 & .853 & .850 & .953  \\
			\hline
			90\% & 5 & noise & .877 & .533 & \textbf{.877} & .873 & .870 & .867 & .863 & .873 & .870 \\
			&  & signal & .967 & .627 & .933 & .833 & \textbf{.940} & .877 & .903 & .923 & .883 \\
			& 5000 & noise & .967 & .613 & .977 & .970 & .973 & .973 & .973 & \textbf{.980}  & .903 \\
			&  & signal & .900 & .493 & .807 & \textbf{.847} & .810 & .810 & .797 & .807  & .813 \\
			\hline
	\end{tabular}}
	\caption{True positive rate (TPR). CDA = complete data analysis; CCA = complete case analysis; MI = feature-level mean imputation; MF = feature-level missForest; mMI = meta-level mean imputation; mMF = meta-level missForest; mPMM = meta-level predictive mean matching, generating $\bm{Z}$ once; cvPMM = meta-level predictive mean matching, generating $\bm{Z}$ five times; MOFA = multi-factor omics analysis imputation. The `best' (highest) values in each row are printed in bold (excluding the values for CDA and CCA since these are not imputation methods).}  \label{tab:tpr}
\end{table}

\begin{table}
	\centering
	\setlength{\tabcolsep}{3pt}
	\resizebox{\textwidth}{!}{
		\begin{tabular}{lllrrrrrrrrr}
			\hline
			missingness & size of missing view & missing view is & CDA & CCA & MI & MF & mMI & mMF & mPMM & cvPMM & MOFA \\
			\hline
			50\% & 5 & noise & .02 & .08 & .04 & .06 & \textbf{.03} & .07 & .07 & .05 & .11 \\
			&  & signal & .05 & .11 & .14 & \textbf{.01} & .19 & .02 & .02 & \textbf{.01} & \textbf{.01} \\
			& 5000 & noise & .04 & .07 & .06 & .24 & \textbf{.05} & .08 & .09 & .09 &.90  \\
			&  & signal & .06 & .05 & .17 & .17 & .20 & \textbf{.05} & .06 & \textbf{.05} &.12 \\
			\hline
			90\% & 5 & noise & .09 & .10 & \textbf{.05} & .16 & .09 & .17 & .17 & .13 & .54 \\
			&  & signal & .02 & .07 & .33 & \textbf{.01} & .32 & .05 & .08 & .10 & .16 \\
			& 5000 & noise & .04 & .11 & \textbf{0} & .43 & .01 & .20 & .11 & .09 & .71 \\
			&  & signal & .07 & .05 & .28 & .22 & .28 & .06 & .07 & \textbf{.04} & \textbf{.04} \\
			\hline
	\end{tabular}}
	\caption{False positive rate (FPR). CDA = complete data analysis; CCA = complete case analysis; MI = feature-level mean imputation; MF = feature-level missForest; mMI = meta-level mean imputation; mMF = meta-level missForest; mPMM = meta-level predictive mean matching, generating $\bm{Z}$ once; cvPMM = meta-level predictive mean matching, generating $\bm{Z}$ five times; MOFA = multi-factor omics analysis imputation. The `best' (lowest) values in each row are printed in bold (excluding the values for CDA and CCA since these are not imputation methods).}  \label{tab:fpr}
\end{table}

\begin{table}
	\centering
	\setlength{\tabcolsep}{3pt}
	\resizebox{\textwidth}{!}{
		\begin{tabular}{lllrrrrrrrrr}
			\hline
			missingness & size of missing view & missing view is & CDA & CCA & MI & MF & mMI & mMF & mPMM & cvPMM & MOFA \\
			\hline
			50\% & 5 & noise & .005 & .022 & .010 & .015 & \textbf{.008} & .018 & .018 & .013 & .030 \\
			&  & signal & .013 & .029 & .035 & \textbf{.003} & .047 & .005 & .006 & \textbf{.003} & \textbf{.003}   \\
			& 5000 & noise & .011 & .018 & .015 & .063 & \textbf{.013} & .020 & .023 & .023 & .236 \\
			&  & signal & .016 & .013 & .044 & .045 & .053 & \textbf{.013} & .016 & \textbf{.013} & .030 \\
			\hline
			90\% & 5 & noise & .023 & .034 & \textbf{.013} & .041 & .024 & .048 & .047 & .035 & .153 \\
			&  & signal & .005 & .022 & .084 & \textbf{.003} & .08 & .013 & .021 & .026 & .044 \\
			& 5000 & noise & .012 & .031 & \textbf{0} & .112 & .003 & .051 & .028 & .023 & .203 \\
			&  & signal & .018 & .032 & .073 & .060 & .073 & .016 & .020 & \textbf{.011} & .012 \\
			\hline
	\end{tabular}}
	\caption{False discovery rate (FDR). CDA = complete data analysis; CCA = complete case analysis; MI = feature-level mean imputation; MF = feature-level missForest; mMI = meta-level mean imputation; mMF = meta-level missForest; mPMM = meta-level predictive mean matching, generating $\bm{Z}$ once; cvPMM = meta-level predictive mean matching, generating $\bm{Z}$ five times; MOFA = multi-factor omics analysis imputation. The `best' (lowest) values in each row are printed in bold (excluding the values for CDA and CCA since these are not imputation methods).}  \label{tab:fdr}
\end{table}

\FloatBarrier

\subsection{Computation Time}

\begin{figure}[h!]
	\centering
	\resizebox{.99\linewidth}{!}{\includegraphics{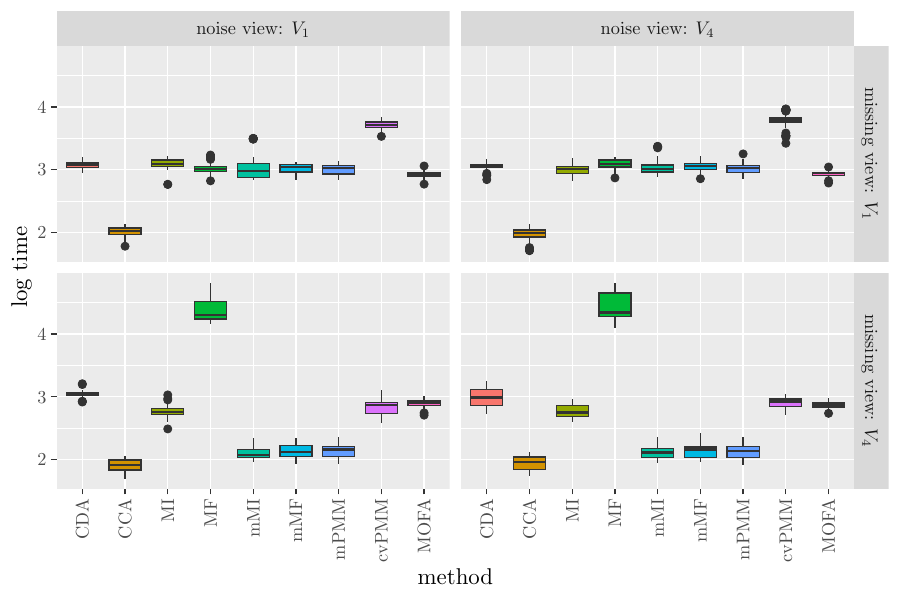}}
	\caption{Logarithm (base 10) of computation time (in seconds + 1) with 90\% missingness. CDA = complete data analysis; CCA = complete case analysis; MI = feature-level mean imputation; MF = feature-level missForest; mMI = meta-level mean imputation; mMF = meta-level missForest; mPMM = meta-level predictive mean matching, generating $\bm{Z}$ once; cvPMM = meta-level predictive mean matching, generating $\bm{Z}$ five times; MOFA = multi-factor omics analysis imputation. $V_1$ is the smallest view, consisting of 5 features. $V_4$ is the largest view, consisting of 5000 features. \label{fig:time_90}}
\end{figure}

\begin{figure}[h!]
	\centering
	\resizebox{.99\linewidth}{!}{\includegraphics{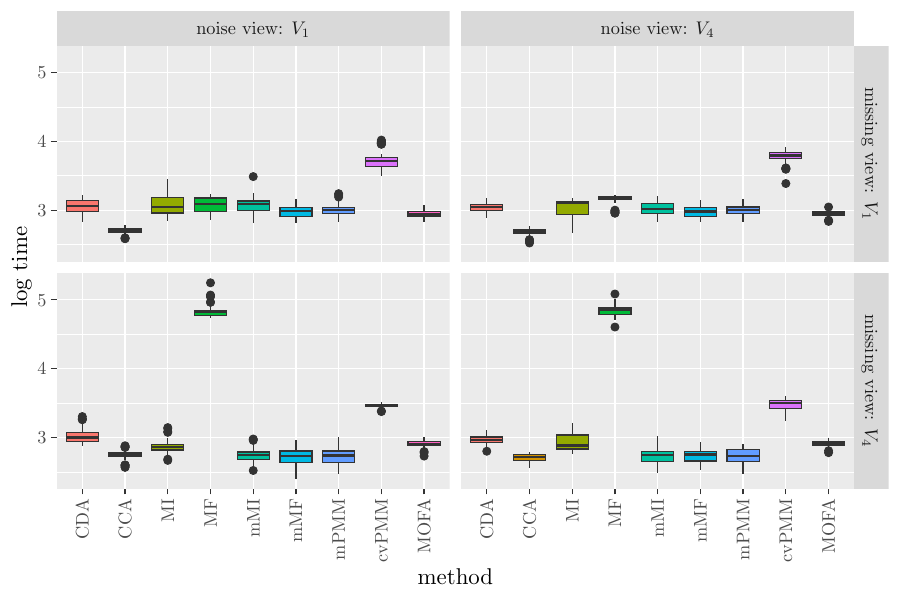}}
	\caption{Logarithm (base 10) of computation time (in seconds + 1) with 50\% missingness. CDA = complete data analysis; CCA = complete case analysis; MI = feature-level mean imputation; MF = feature-level missForest; mMI = meta-level mean imputation; mMF = meta-level missForest; mPMM = meta-level predictive mean matching, generating $\bm{Z}$ once; cvPMM = meta-level predictive mean matching, generating $\bm{Z}$ five times; MOFA = multi-factor omics analysis imputation. $V_1$ is the smallest view, consisting of 5 features. $V_4$ is the largest view, consisting of 5000 features. \label{fig:time_50}}
\end{figure}

Figure \ref{fig:time_90} shows the log computation time for all imputation methods, as well as CDA and CCA, with 90\% missingness. Figure \ref{fig:time_50} shows the log computation time with 50\% missingness. Note that the computation times reported here capture both imputation and model fitting. It can be observed that CCA has the lowest computation time across all plots, since it constitutes a reduction in the amount of observations in the training data. By comparing MI with mMI, and MF with mMF, it can be observed that meta-level imputation is faster than base-level imputation. This is particularly noticeable when the view to be imputed is large (i.e., when $V_4$ is missing), when MF is the slowest method by a large margin. cvPMM is considerably slower than mPMM, and is the slowest method when the view to be imputed is small (i.e., when $V_1$ is missing). 

\subsection{Summary of Simulation Results}

Overall, we can summarize the results of our simulations as follows:

\begin{itemize}
	\item Meta-level imputation is much faster than feature-level imputation.
	\item Meta-level imputation performs approximately as well as feature-level imputation in terms of accuracy, TPR, FPR, and FDR.
	\item Mean imputation is very fast and sometimes works very well, but its performance heavily depends on whether the missing values correspond to signal or noise.
	\item MOFA-based imputation does not perform very well.
\end{itemize}

\FloatBarrier

\section{Application to multi-view MRI data}

We apply the imputation methods used in our simulation study to a real multi-view data set which was collected as part of the Prospective Registry on Dementia (PRODEM) \citep{PRODEM} and the Austrian Stroke Prevention Study (ASPS) \citep{ASPS1,ASPS2}. This data set consists of 40 views in the form of magnetic resonance imaging (MRI) measures obtained from 3 different scan types, namely structural MRI, diffusion-weighted MRI and resting-state functional MRI. The outcome variable is whether or not a subject has Alzheimer's Disease (AD). The data set contains 249 subjects, of which 76 were clinically diagnosed with probable AD, and the other 173 are cognitively normal elderly controls. For a more detailed description of the data, and an application of StaPLR to the complete data set, we refer to \citet{StaPLR3}. Note that one could also take the 3 different scan types as the views, or fit a hierarchical model with the MRI measures nested within scan types \citep{StaPLR3}. However, for the purpose of this demonstration we will use the MRI measures, since this leads to the largest number of views. An overview of these views can be observed in Table \ref{tab:features}. \par 
We introduce missingness in the data by setting either the views obtained from the structural scan, or those obtained from the functional scan, to be missing for 75\% of the subjects. We choose 75\% as the percentage of missingness based on the observation of \citet{Thung2018} that in a set of three different views from the Alzheimer's Disease Neuroimaging Initiative (ADNI)\citep{ADNI} database, only 25\% of observations were complete \citep{Thung2018}. We choose to set either the structural views or the functional views to be missing, since they provide two conditions which are challenging in different ways. Although in real data the `true' predictive views are unknown, it has previously been observed that the structural views, in particular grey matter density and cortical thickness, contain strong signal for separating Alzheimer's patients from healthy controls, and these views tend to dominate the classification model \citep{StaPLR3}. The functional views, on the other hand, are often very noisy, and although they can be relevant to the classification, their added value compared with that of grey matter density and cortical thickness is often limited \citep{StaPLR3}. Thus, removing the structural views removes a small set of strongly predictive views (5 views, or 266 features) from the data, requiring them to be imputed from a much larger set of much more noisy views, which is a challenging imputation problem. Imputation of the functional views is also challenging, not because of the removal of strong signal, but because of the sheer volume of missing data: This corresponds to 31 missing views or 2,876,169 features that have to be imputed. \par
We evaluate the performance of the different imputation methods by calculating the classification accuracy and Brier score using 10 repeats of 10-fold cross-validation. Note that when an observation is included in the test partition it does not have missing values; only observations used for training have missing values. \par
\begin{figure}[h!]
	\centering
	\resizebox{.99\linewidth}{!}{\includegraphics{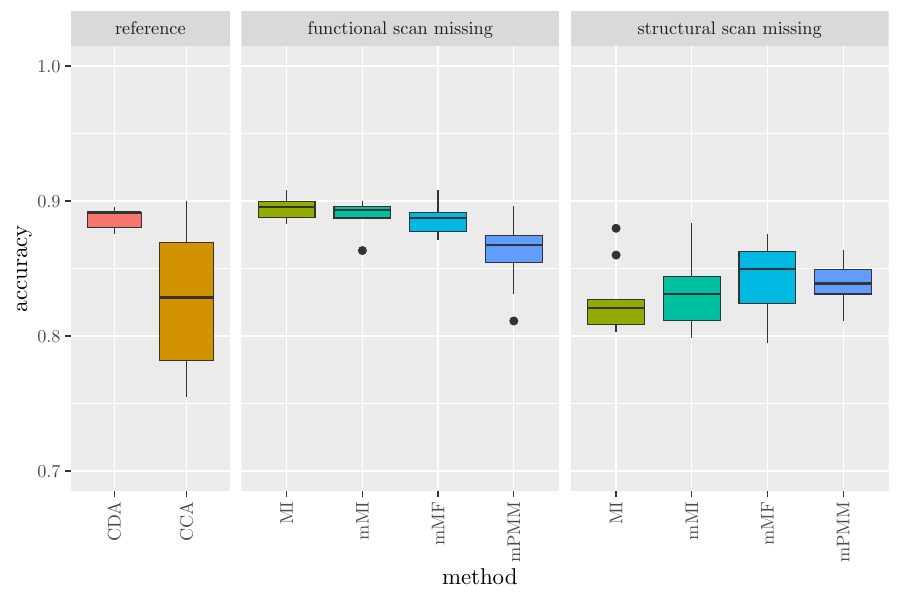}}
	\caption{Classification accuracy on the multi-view MRI data obtained from 10 repeats of 10-fold cross-validation, with either the structural or functional scan set to missing for 75\% of the observations. CDA = complete data analysis; CCA = complete case analysis; MI = feature-level mean imputation; mMI = meta-level mean imputation; mMF = meta-level missForest; mPMM = meta-level predictive mean matching, generating $\bm{Z}$ once. \label{fig:mri_accuracy}}
\end{figure}
\begin{figure}[h!]
	\centering
	\resizebox{.99\linewidth}{!}{\includegraphics{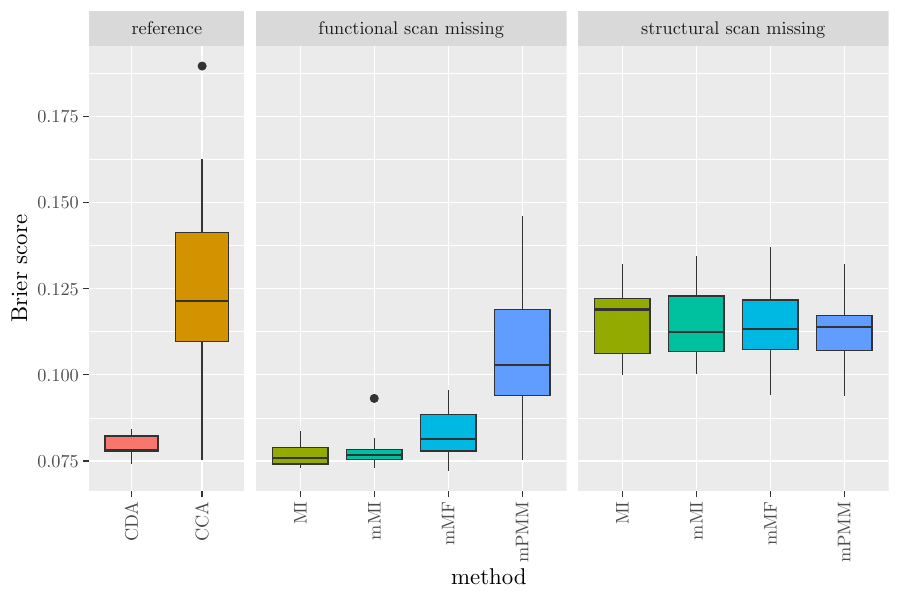}}
	\caption{Brier score on the multi-view MRI data obtained from 10 repeats of 10-fold cross-validation, with either the structural or functional scan set to missing for 75\% of the observations. CDA = complete data analysis; CCA = complete case analysis; MI = feature-level mean imputation; mMI = meta-level mean imputation; mMF = meta-level missForest; mPMM = meta-level predictive mean matching, generating $\bm{Z}$ once. \label{fig:mri_brier}}
\end{figure}

The results can be observed in Figure \ref{fig:mri_accuracy} and Figure \ref{fig:mri_brier}. For the meta-level imputation using PMM, we only included the variant where we calculate the matrix of cross-validated predictions once (mPMM). No results were obtained for feature-level imputation using PMM, missForest, or MOFA since all of these methods turned out to be computationally infeasible for this data because they required more than the maximum amount of RAM we had available per process (64 GB). However, even if more RAM had been available, it is unlikely that PMM or missForest imputation at the feature level would have been completable within any reasonable time-frame: PMM was already infeasible for our simulation study, and feature-level missForest could not complete a single iteration within 10 days of computation before running out of memory. \par
\begin{figure}[h!]
	\centering
	\resizebox{.99\linewidth}{!}{\includegraphics{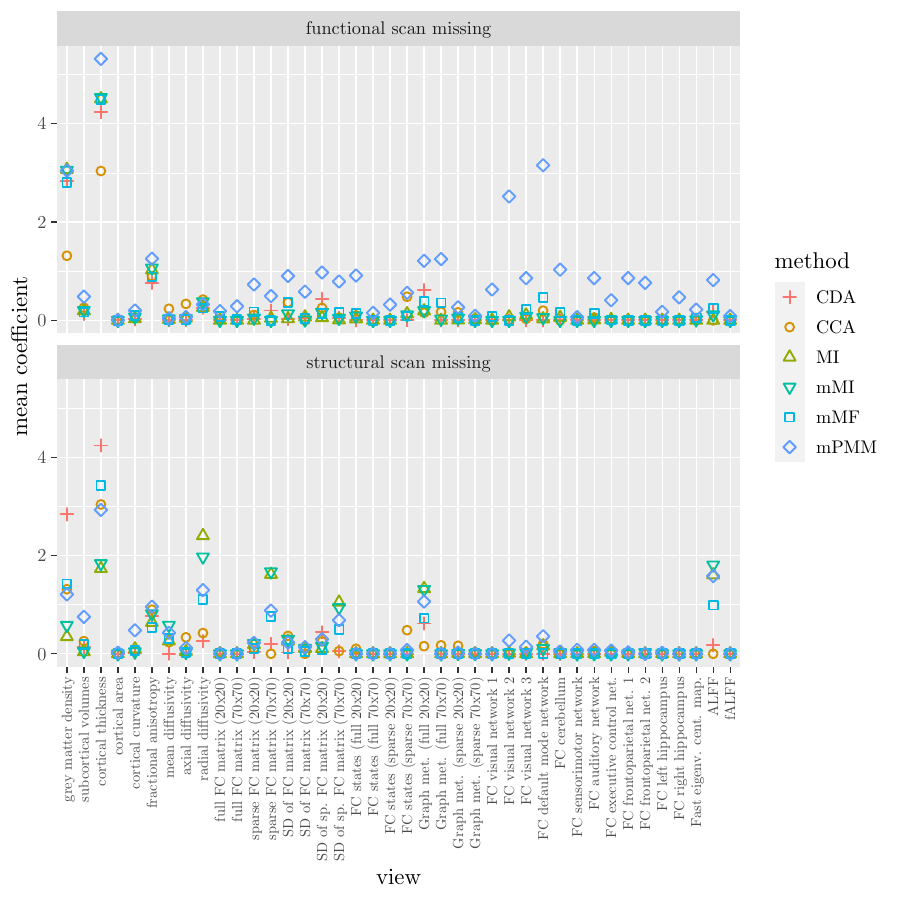}}
	\caption{Mean view-level coefficients obtained from 10 repeats of 10-fold cross-validation. Of the different imputation methods, meta-level missForest imputation (blue squares) produces coefficients which are on average closest to those obtained from training the model on the complete data (red crosses), with a mean squared error (MSE) of 0.587 when the structural scan is missing, and an MSE of 0.618 when the functional scan is missing. CDA = complete data analysis; CCA = complete case analysis; MI = feature-level mean imputation; mMI = meta-level mean imputation; mMF = meta-level missForest; mPMM = meta-level predictive mean matching, generating $\bm{Z}$ once. \label{fig:mri_coefficients}}
\end{figure}

\begin{figure}[h!]
	\centering
	\resizebox{.99\linewidth}{!}{\includegraphics{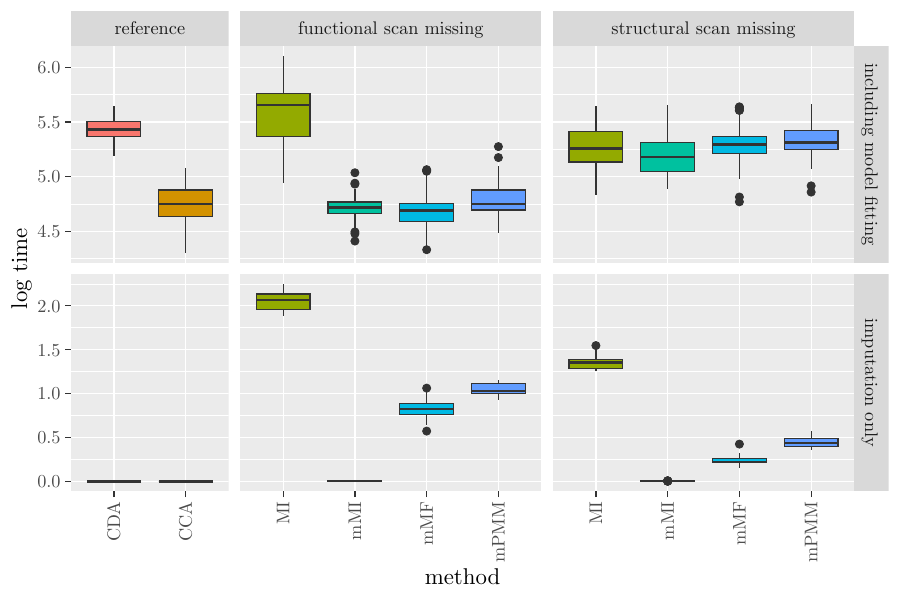}}
	\caption{Logarithm (base 10) of computation time (in seconds + 1) on the multi-view MRI data. The top row displays the computation time including the time used for fitting the model; the bottom row displays the computation time of the imputation only. Either the structural or functional scan was missing for 75\% of the observations. CDA = complete data analysis; CCA = complete case analysis; MI = feature-level mean imputation; mMI = meta-level mean imputation; mMF = meta-level missForest; mPMM = meta-level predictive mean matching, generating $\bm{Z}$ once. \label{fig:mri_time}}
\end{figure}

 
It can be observed in Figures \ref{fig:mri_accuracy} and \ref{fig:mri_brier} that mean imputation, either at the base level or the meta level, performs very well if the functional scan is missing, even performing slightly better than the complete data. However, it performs poorly when the structural scan is missing. Meta-level missForest imputation performs better than meta-level PMM, especially when the functional scan is missing.
In real data the `true' underlying model is unknown and quantities such as the TPR, FPR and FDR cannot be calculated. However, we can compare the coefficients obtained from fitting the model on the imputed data with those obtained from fitting the model on the complete data (Figure \ref{fig:mri_coefficients}). Again it can be observed that mean imputation performs poorly when the structural scan is missing, as it attenuates the coefficients of the most important views (grey matter density and cortical thickness) while inflating those of comparatively less important views (such as radial diffusivity and ALFF). This behavior is comparable to that observed in our simulation study (Section \ref{sect:view_selection}), although in the real data it does result in a substantial reduction in accuracy (Figure \ref{fig:mri_accuracy}). Using meta-level PMM when the functional scan is missing leads to many additional functional views being selected, which is also associated with a reduction in accuracy (Figure \ref{fig:mri_accuracy}). Out of all the imputation methods, meta-level missForest imputation leads to coefficients which are on average closest to those obtained from the complete data (Figure \ref{fig:mri_coefficients}). In addition, it can be observed in Figure \ref{fig:mri_time} that it is also faster than meta-level PMM. 

\begin{table}
	\centering
	\resizebox{\textwidth}{!}{
		\begin{tabular}{llr}
			\hline
			scan type & MRI measure (view) & number of features \\
			\hline
			structural MRI & (1) grey matter density & 48 \\
			& (2) subcortical volumes & 14 \\
			& (3) cortical thickness & 68 \\
			& (4) cortical area & 68 \\
			& (5) cortical curvature & 68 \\
			\hline
			diffusion MRI & (6) fractional anisotropy & 20 \\
			& (7) mean diffusivity & 20 \\
			& (8) axial diffusivity & 20 \\
			& (9) radial diffusivity & 20 \\
			\hline
			resting state fMRI & (10) full FC correlation matrix (20 $\times$ 20) & 190 \\
			& (11) full FC correlation matrix (70 $\times$ 70) & 2,415 \\
			& (12) sparse partial FC correlation matrix (20 $\times$ 20) & 189 \\
			& (13) sparse partial FC correlation matrix (70 $\times$ 70) & 2,337 \\
			& (14) SD of full FC matrix (20 $\times$ 20) & 190 \\
			& (15) SD of full FC matrix (70 $\times$ 70) & 2,415 \\
			& (16) SD of sparse partial FC matrix (20 $\times$ 20) & 190 \\
			& (17) SD of sparse partial FC matrix (70 $\times$ 70) & 2,414 \\
			& (18) FC states of full FC matrix (20 $\times$ 20) & 5 \\
			& (19) FC states of full FC matrix (70 $\times$ 70) & 5 \\
			& (20) FC states of sparse partial FC matrix (20 $\times$ 20) & 5 \\
			& (21) FC states of sparse partial FC matrix (70 $\times$ 70) & 5 \\
			& (22) Graph metrics of full FC matrix (20 $\times$ 20) & 124 \\
			& (23) Graph metrics of full FC matrix (70 $\times$ 70) & 424 \\
			& (24) Graph metrics of sp. par. FC matrix (20 $\times$ 20) & 123 \\
			& (25) Graph metrics of sp. par. FC matrix (70 $\times$ 70) & 423 \\
			& (26) FC with visual network 1 & 190,981 \\
			& (27) FC with visual network 2 & 190,981 \\
			& (28) FC with visual network 3 & 190,981 \\
			& (29) FC with default mode network & 190,981 \\
			& (30) FC with the cerebellum & 190,981 \\
			& (31) FC with sensorimotor network & 190,981 \\
			& (32) FC with auditory network & 190,981 \\
			& (33) FC with executive control network & 190,981 \\
			& (34) FC with frontoparietal network 1 & 190,981 \\
			& (35) FC with frontoparietal network 2 & 190,981 \\
			& (36) FC with left hippocampus & 190,981 \\
			& (37) FC with right hippocampus & 190,981 \\
			& (38) Fast eigenvector centrality mapping & 190,981 \\
			& (39) ALFF & 190,981 \\
			& (40) fALFF & 190,981 \\
			\hline
			total & & 2,876,515\\
			\hline
	\end{tabular}}
	\caption{Overview of the different views and corresponding number of features. SD = standard deviation; FC = functional connectivity; ALFF = amplitude of low frequency fluctuations, fALFF = fractional ALFF. Table adapted from \citet{StaPLR3}.}  \label{tab:features}
\end{table}

\FloatBarrier

\section{Discussion}

We proposed a meta-level imputation method for missing values in multi-view data. We evaluated the proposed method with several imputation algorithms using simulations and an application to real multi-view MRI data. We compared the meta-level implementation of these imputation algorithms with their feature-level implementation whenever possible, and also compared the results with those of a dedicated multi-view imputation method (MOFA), as well as complete case analysis and complete data analysis. In line with \citet[p. 697]{Orchard1972}, who stated that ``the best way to treat missing data is not to have them", we found that complete data analysis often performs best. Often, but not always: In a real multi-view data set, if we removed 75\% of observed values for 31 out of the 40 views and imputed them using the unconditional mean, predictive accuracy actually improved slightly. At a first glance it seems counter-intuitive that removing large volumes of data would improve performance. However, recall that MI reduces the variance of imputed features and attenuates the correlation with the outcome \citep{Stef2018}. As such, mean imputation is a kind of regularization, applied only to the views in which missing values occur. If a view is very noisy, this regularization can be beneficial, but if the view contains strong signal, this regularization can be detrimental. Note, however, that since we use models that already contain explicit regularization terms, the exact effect on the performance of the model will depend on the interplay between the view-specific regularization caused by mean imputation and the overall regularization through the penalty parameters. Nevertheless, it is clear that the performance of mean imputation heavily depends on whether the missing data corresponds to signal or to noise. Obviously, whether a view corresponds to signal or to noise is generally unknown beforehand. 
Thus, unconditional mean imputation cannot be recommended as a general imputation method, and is especially unsuitable if the goal is not just to predict, but also to perform view selection. \par

Our results show that meta-level imputation is much faster than feature-level imputation, at no or very minor costs to predictive and view-selection  performance. In fact, meta-level imputation allows the use of sophisticated imputation algorithms such as predictive mean matching (PMM) or missForest (MF) which are computationally heavy or infeasible when applied at the feature level. In our experiments meta-level imputation using PMM or MF performed better than MOFA-based imputation. \par 
For the meta-level implementation of PMM, we used two different ways of combining the imputed data sets. The first one is generating the $\bm{Z}$ matrix once based on observed data, imputing it 5 times, then training the meta-learner on the averaged data (mPMM). The second is generating the $\bm{Z}$ matrix 5 times, imputing each of these matrices once, and then training the meta-learner on the averaged data (cvPMM). The cvPMM approach is much more computationally expensive (Figures \ref{fig:time_50} and \ref{fig:time_90}), but also averages over different allocations to the cross-validation folds, instead of just over different imputations. However, in our simulations, this approach did not result in an improvement in classification performance, and only small differences in view selection performance. In our real data application we only applied the mPMM variant, which performed considerably worse than mMF. It is possible that cvPMM would have performed better, but given the results of our simulations we found this to be unlikely and not worth the additional computation cost. Of course, other methods of combining imputations are possible. For example, one could train the meta learner separately on each imputated data set, leading to 5 separate StaPLR models. However, this leads to additional questions on how to combine the separate models, and given the observed differences between mPMM and cvPMM, we would not expect this approach to cause large gains in performance. \par 
Using missForest in our simulations, we found that feature-level MF sometimes performed better than mMF or mPMM/cvPMM in terms of MSEP, but these observed differences in terms of predicted probabilities did not affect classification performance. In our real data analysis, mMF was also the best-performing imputation method in terms of classification accuracy (excluding the aforementioned use of mean imputation for the functional views). For view selection, the observed differences between MF and mMF/mPMM/cvPMM are not entirely consistent. However, it is worth noting that in some cases base level MF resulted in considerably higher values of FPR and FDR than mMF. Overall, the meta-level implementation of missForest does not perform worse than the base-level implementation in most important metrics. Additionally, in the real data the coefficients obtained from training the meta-learner on the values imputed by mMF were the closest to those obtained from the complete data. Thus, out of all the imputation methods for multi-view data that we considered in this study, we would most recommend meta-level missForest imputation because it consistently performed well in both the simulations and real data application, and is much faster than the feature-level implementation. \par

Throughout this article we have assumed that while the number of features $m$ was larger than $n$, the number of views $V$ was always smaller. It is natural then to wonder how the meta-level imputation methods would perform if $n < V$ (i.e., if $\bm{Z}$ were high-dimensional). Given that missForest can handle high-dimensional data, as is evident by the performance of MF in this article, mMF could most likely be used without issue. The same cannot be said for mPMM or cvPMM, since standard PMM based on the linear model struggles even with small high-dimensional data sets (see \ref{sect:pmm}). However, alternative approaches could be constructed. For example, an imputation method based on the lasso \citep{lasso_imputation_1, lasso_imputation_2} is included in \texttt{mice} \citep{mice}, and appeared to work quite well for a small high-dimensional data set such as the one described in \ref{sect:pmm} (results not shown), although the associated computation time still made it infeasible to include as a base-level imputation method in this study. Nevertheless, since generally $V \ll m$, it has potential as a meta-level imputation method in cases where $\bm{Z}$ is high-dimensional. \par 
In our simulations missingness occurred only in the training set, but in practical settings it is likely that missingness occurs also in the test set. In this case, \citet{Josse2019} recommend that the same imputation model that was used to impute the training set is used to impute the test set. Unfortunately, this is not easy to do in practice, since many implementations of imputation methods, including \texttt{mice} \citep{mice} and \texttt{missForest} \citep{missForest}, effectively work as `black boxes' that do not separate the imputation model from its use to complete the data \citep{Josse2019}. A possible solution that has been suggested for this problem is to take a semi-supervised approach \citep{Josse2019}, jointly imputing the training and test set \citep{Kapelner2015}, and then learning the prediction model on the training set only, but this does require that the training and test set are available at the same time \citep{Josse2019}.   \par 
Throughout most of this article we assumed that the data was MCAR, and gave examples of why this is often a realistic assumption in the case of multi-view missing data. Nevertheless, in practice there will also be cases where the data is MAR or even MNAR. Multiple imputation methods such as MF and PMM are known to provide unbiased parameter estimates even under MAR. Of course, this is assuming the learning algorithm generates unbiased estimates in general, which is not the case for StaPLR, where the exact values of the parameter estimates are generally not of interest. How the theoretical knowledge about multiple imputation methods at the feature-level translates to imputation at the meta-level under MAR or MNAR, and how it affects view selection and classification accuracy in these settings is a topic for future research. \par 
This article focuses on the imputation of missing values. As previously stated in Section \ref{sect:methods}, other methods for handling missing data exist which do not explicitly impute missing values, including likelihood-based methods such as full information maximum likelihood (FIML) \citep{Arbuckle1996, Myrtveit2001}. Such methods may provide an elegant way for dealing with missing values in models based on maximum likelihood estimation, or extensions thereof such as multi-view (deep) Gaussian process modeling \citep{Sun2021, Dong2023}. However, such methods can have numerous computational difficulties since the resulting optimization problem typically has no closed-form solution and may have many local optima \citep{Little2020, Smola2005}. Multiple imputation methods are more broadly applicable \citep{Stef2018, Josse2019, Myrtveit2001}, and have the added advantage of making the imputed values explicit, allowing them to be inspected and analyzed \citep{Stef2018}. \par  
In summary, we have proposed a meta-level imputation method for missing values in multi-view data. We have evaluated the method with several imputation algorithms using simulations and an application to real multi-view MRI data. The results show that meta-level imputation produces competitive results in terms of test accuracy and view selection at a much lower computational cost, and makes the use of advanced multiple imputation methods such as missForest and predictive mean matching possible in high-dimensional settings where they were previously computationally infeasible. Meta-level imputation using missForest performed particularly well compared with both its feature-level implementation, and with meta-level imputation using predictive mean matching. Although we used the StaPLR algorithm specifically, the proposed imputation method can be applied to any form of multi-view stacking.

\section{Declaration of Interest}

This research was funded by Leiden University. The authors declare no conflicts of interest.

\section{Data Availability Statement}

Relevant R code is included in the supplementary materials. The MRI data used in this study is available upon direct request to the 5th author (R. Schmidt); a formal data sharing agreement is mandatory. 

\newpage
\appendix

\section{Infeasibility of Predictive Mean Matching Using MICE}\label{sect:pmm}

\FloatBarrier

We first show empirically how predictive mean matching using \texttt{mice} leads to imputations of poor quality when the data is high-dimensional. We generate a high-dimensional, but comparatively small single-view data set with $n= 200$ independent observations of $m = 400$ normally distributed features and one binary outcome $y$. Only the first 50 values of the first feature are missing. Compared to the main simulations presented in this paper, this is a very easy missing data problem.\par
\begin{figure}[h!]
	\centering
	\resizebox{.9\linewidth}{!}{\includegraphics{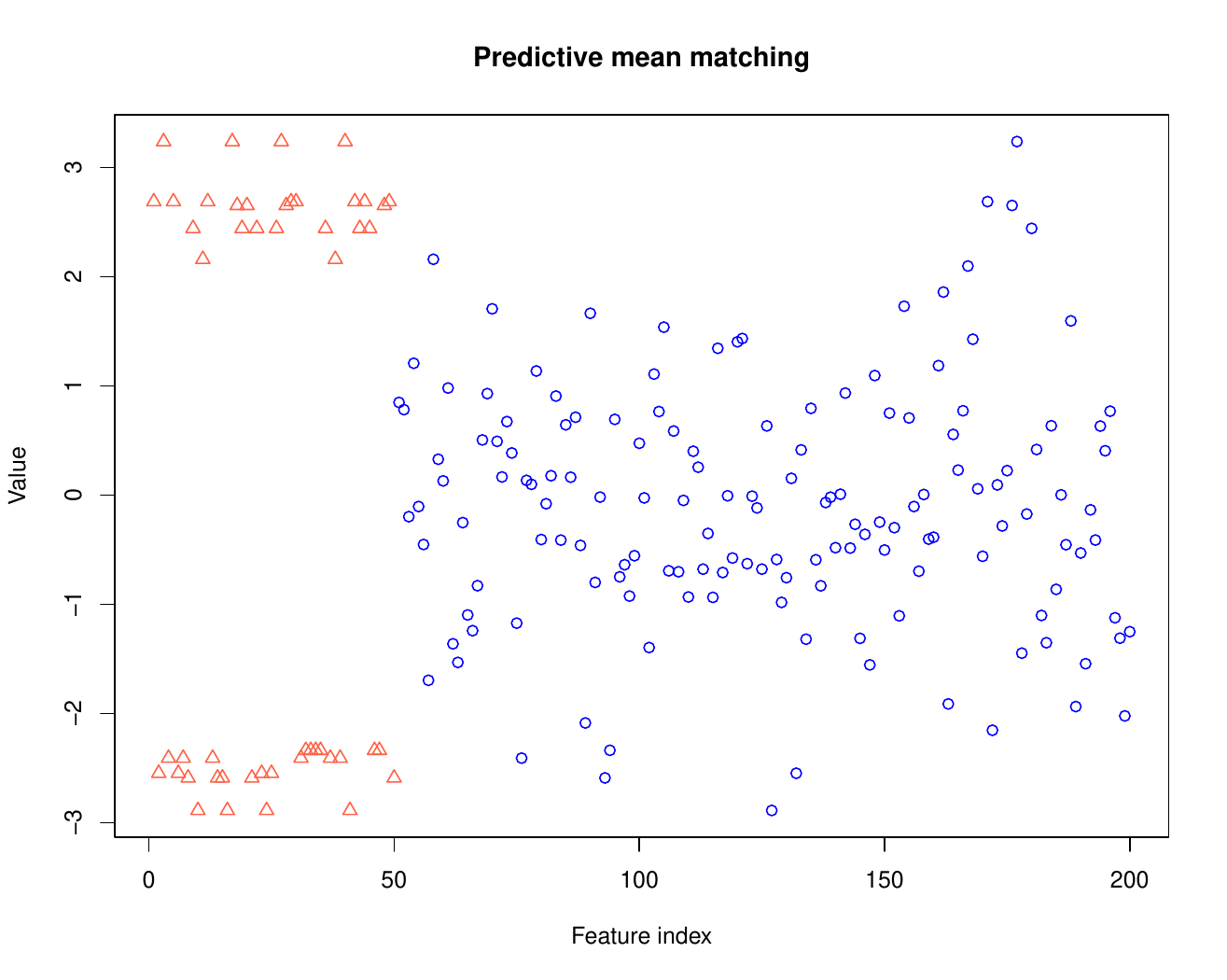}}
	\caption{Imputed (red triangles) and observed (blue circles) values of feature 1 after imputing the missing values with PMM using \texttt{mice}. \label{fig:pmm_failure}}
\end{figure}

 
We generate an imputed data set with PMM using \texttt{mice}. Figure \ref{fig:pmm_failure} shows that the imputed values (first 50 observations; triangles) of feature one are generally at the extreme end of the distribution of observed values, and their distribution is non-normal and very different from that of the observed values (last 150 observations; circles). R code to replicate this image is included in the supplementary materials. The likely cause of this behavior is that the linear model is not well-behaved under high-dimensionality, meaning some form of (ridge) regularization is required \citep{Stef2018, mice}, but the amount of regularization required to find a `good' solution is not obvious. Recommended values for the penalty parameter in the range $10^{-6}$ to $10^{-4}$ do not produce good results, even in this relatively simple setting. \par
Even when good results could be obtained, two practical problems make calculations using \texttt{mice} infeasible for large high-dimensional data sets. The first, and most important one, is time: Even for the simplest setting included in our experiments, a single iteration of a single imputation of a single variable took well over 24 hours to compute. The second, which was not directly relevant to our simulation study but will be for larger problems, is memory handling: A test run of PMM with \texttt{mice} on a data set consisting of 200 observations of 75,000 variables (120 MB of data), which is small by medical imaging standards, already required 42 GB of RAM. 

\FloatBarrier

\section{Binomial Deviance} \label{sect:bin_dev}

\FloatBarrier

The binomial deviance is defined as:
\begin{equation*}
    D = -2\sum_{i=1}^{n} y_i \log(\hat{p}_i) + (1-y_i)\log(1 - \hat{p}_i),
\end{equation*}
where $y_i$ is the observed class label for test set observation $i$, and $\hat{p}_i$ is the predicted probability of belonging to class 1.

\begin{figure}[h!]
	\centering
	\resizebox{.9\linewidth}{!}{\includegraphics{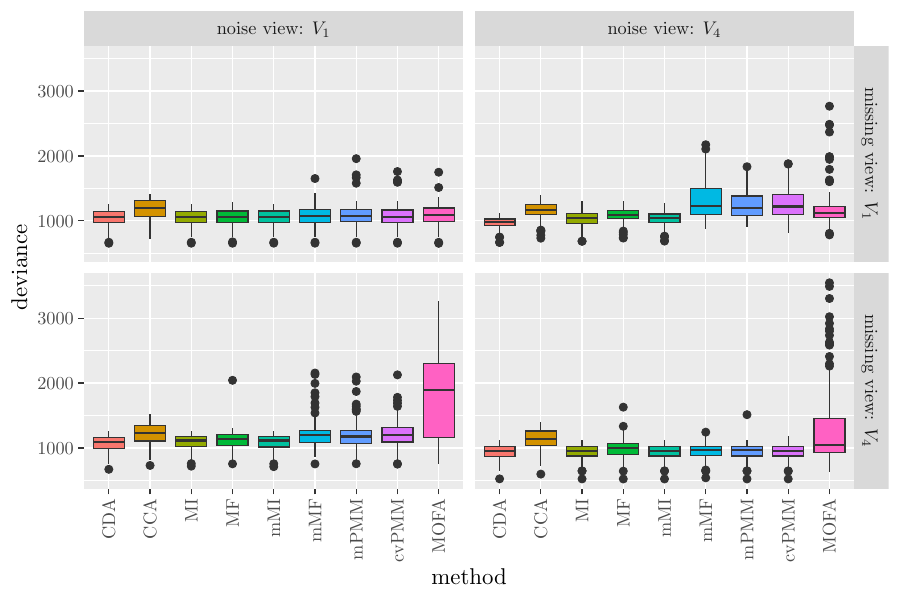}}
	\caption{Binomial deviance with 90\% missingness. CDA = complete data analysis; CCA = complete case analysis; MI = feature-level mean imputation; MF = feature-level missForest; mMI = meta-level mean imputation; mMF = meta-level missForest; mPMM = meta-level predictive mean matching, generating $\bm{Z}$ once; cvPMM = meta-level predictive mean matching, generating $\bm{Z}$ five times; MOFA = multi-factor omics analysis imputation. $V_1$ is the smallest view, consisting of 5 features. $V_4$ is the largest view, consisting of 5000 features. \label{fig:dev_90}}
\end{figure}

\begin{figure}[h!]
	\centering
	\resizebox{.9\linewidth}{!}{\includegraphics{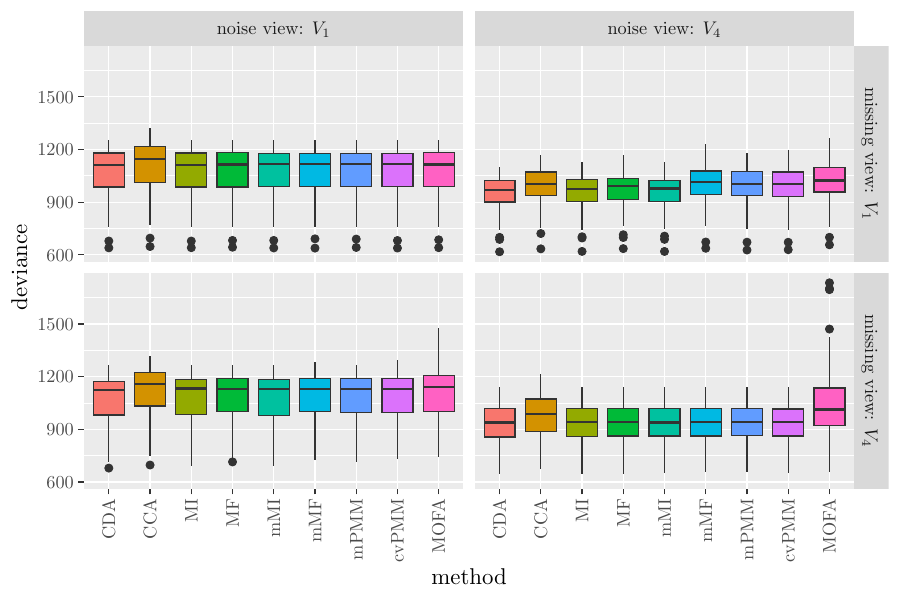}}
	\caption{Binomial deviance with 50\% missingness. CDA = complete data analysis; CCA = complete case analysis; MI = feature-level mean imputation; MF = feature-level missForest; mMI = meta-level mean imputation; mMF = meta-level missForest; mPMM = meta-level predictive mean matching, generating $\bm{Z}$ once; cvPMM = meta-level predictive mean matching, generating $\bm{Z}$ five times; MOFA = multi-factor omics analysis imputation. $V_1$ is the smallest view, consisting of 5 features. $V_4$ is the largest view, consisting of 5000 features. \label{fig:dev_50}}
\end{figure}

\FloatBarrier

\bibliographystyle{IEEEtranN}
\bibliography{Bibliography} 

\end{document}